%% file: main.tex
\documentclass[10pt,twocolumn,letterpaper]{article}
\usepackage{authblk}

\usepackage{iccv}              

\input{preamble}

\usepackage{bm}
\usepackage{multirow}
\usepackage{makecell}
\usepackage{colortbl}

\definecolor{iccvblue}{rgb}{0.21,0.49,0.74}
\usepackage[pagebackref,breaklinks,colorlinks,allcolors=iccvblue]{hyperref}

\def\paperID{7305} 
\def\confName{ICCV}
\def\confYear{2025}

\begin{document}

\title{RealCam-I2V: Real-World Image-to-Video Generation with \\ Interactive Complex Camera Control}

\author[1,2*]{Teng Li}
\author[1,2*]{Guangcong Zheng}
\author[1,2]{Rui Jiang}
\author[1]{Shuigen Zhan}
\author[1]{Tao Wu}
\author[1]{Yehao Lu}
\author[3]{Yining Lin}
\author[2]{Chuanyun Deng}
\author[2]{Yepan Xiong}
\author[2]{Min Chen}
\author[2]{Lin Cheng}
\author[1\dag]{Xi Li}

\affil[1]{
College of Computer Science \& Technology, 
Zhejiang University
}
\affil[2]{
Central Media Technology Institute, 
2012 Lab, 
Huawei
}
\affil[3]{
Supremind
}

\twocolumn[{%
\renewcommand\twocolumn[1][]{#1}%
\maketitle
\vspace{-9mm}

\begin{center}
    \centering
    \captionsetup{type=figure}
    \includegraphics[width=0.96\textwidth]{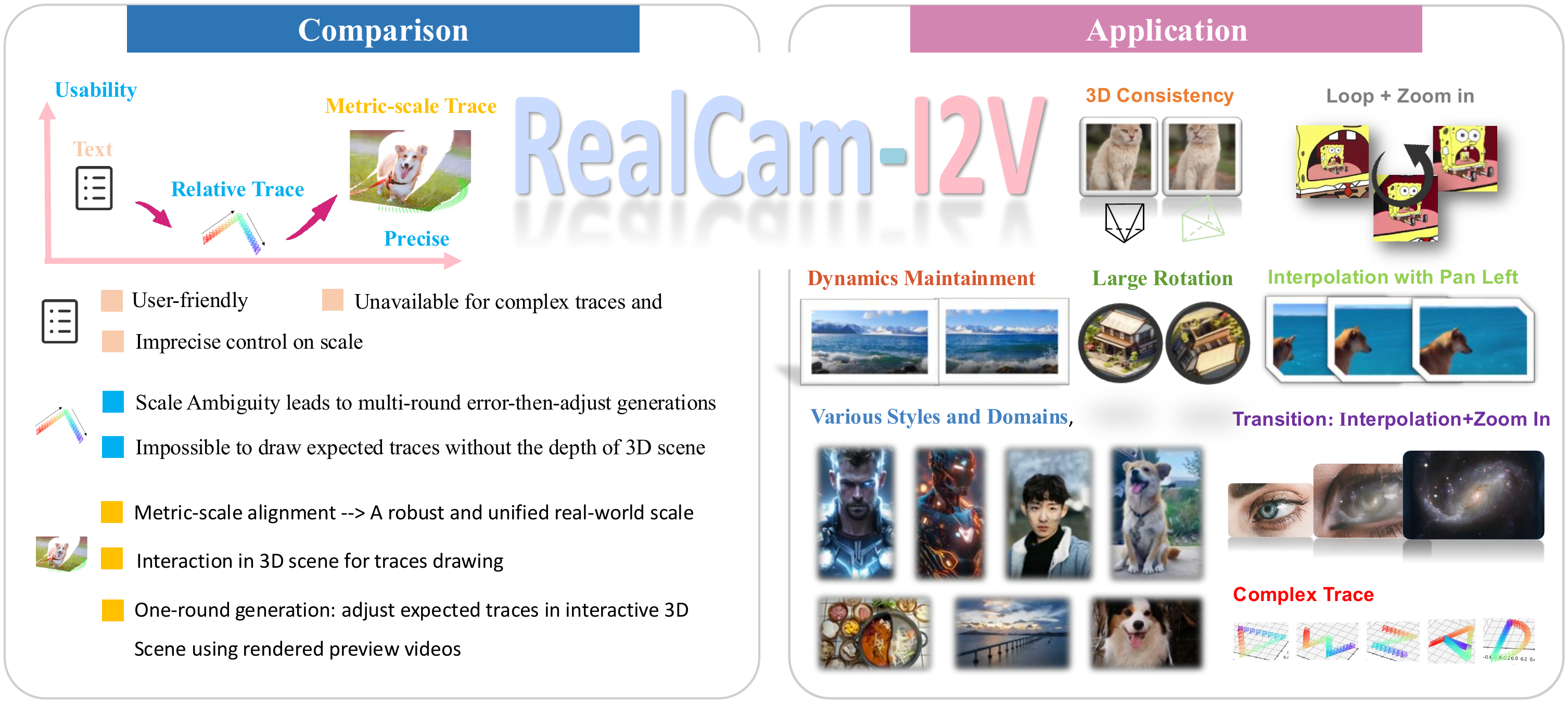}
    \vspace{-2mm}
    \caption{We propose RealCam-I2V, a camera controllable image-to-video generation framework for complex real-world camera control and extra applications including camera-controlled loop video generation, generative frame interpolation, and smooth scene transitions.}
    \label{first_img}
\end{center}
}]

\maketitle

\renewcommand{\thefootnote}{\fnsymbol{footnote}}
\footnotetext[1]{Equal Contribution}

\input{sec/0_abstract}
\input{sec/1_intro}

\input{sec/2_related_work}

\input{sec/3_method}

\input{sec/4_experiment}
\input{sec/5_conclusion}
\clearpage

{
    \small
    \bibliographystyle{ieeenat_fullname}
    \bibliography{main}
}
\clearpage

\input{sec/supplementary}

\end{document}

%% file: preamble.tex
\newcommand{\red}[1]{{\color{red}#1}}


%% file: sec/0_abstract.tex
\begin{abstract}
Recent advancements in camera-trajectory-guided image-to-video generation offer higher precision and better support for complex camera control compared to text-based approaches. 
However, they also introduce significant usability challenges, as users often struggle to provide precise camera parameters when working with arbitrary real-world images without knowledge of their depth nor scene scale.
To address these real-world application issues, we propose RealCam-I2V, a novel diffusion-based video generation framework that integrates monocular metric depth estimation to establish 3D scene reconstruction in a preprocessing step. 
During training, the reconstructed 3D scene enables scaling camera parameters from relative to metric scales, ensuring compatibility and scale consistency across diverse real-world images. 
In inference, RealCam-I2V offers an intuitive interface where users can precisely draw camera trajectories by dragging within the 3D scene.
To further enhance precise camera control and scene consistency, we propose scene-constrained noise shaping, which shapes high-level noise and also allows the framework to maintain dynamic and coherent video generation in lower noise stages.
RealCam-I2V achieves significant improvements in controllability and video quality on the RealEstate10K and out-of-domain images. We further enables applications like camera-controlled looping video generation and generative frame interpolation. 
Project page: \url{zgctroy.github.io/RealCam-I2V}.
\end{abstract}

%% file: sec/1_intro.tex
\vspace{-2mm}
\section{Introduction}
\label{Introduction}
Recent advancements in image-to-video generation \cite{Guo2023, he2022latent, chen2023videocrafter1, chen2024videocrafter2,  yang2024cogvideox} have significantly improved controllability over synthesized videos.
However, challenges remain in achieving realistic, controllable camera movement within complex real-world scenes.
Text-based camera-control methods \cite{Guo2023, Blattmann2023, li2024image, hu2024motionmaster, jain2024peekaboo, wang2024videocomposer}, like traditional diffusion-based video generation, are intuitive and straightforward but lack precision in explicit control over camera parameters, such as angle, scale, and movement direction.
This limitation has spurred the development of camera-trajectory-guided approaches, which attempt to address these issues by offering finer control over camera movement.


Current camera-trajectory-guided methods typically rely on relative camera trajectories, as seen in models like MotionCtrl  \cite{Wang2024Motionctrl}, CameraCtrl \cite{He2024Cameractrl}, CamCo \cite{xu2024camco}, and CamI2V \cite{zheng2024cami2v}. While these approaches provide more control than text-based models, they are fundamentally limited by their reliance on relative scale trajectories. Training on relative scales results in inconsistencies when applied to real-world scenes, where metric scale is crucial for realistic depth perception. Additionally, without access to depth information, users find it challenging to draw precise trajectories, making these methods difficult to use effectively.

To overcome these limitations, we propose RealCam-I2V, a video generation framework that integrates monocular depth estimation as a preprocessing step to construct a robust, metric-scale 3D scene. Our approach leverages the Depth Anything v2~\cite{depth_anything_v2} model (metric version) to predict the metric depth of a user-provided reference image, reprojecting its pixels back into camera space to create a stable 3D representation. This 3D scene serves as the foundation for camera control, providing a consistent and metric scale that is critical for real-world applications.

In the training stage, we align the reconstructed 3D scene of the reference image with the point cloud of each video sample, reconstructed using COLMAP \cite{Schonberger2016}, a structure-from-motion (SfM) method. This alignment allows us to rescale COLMAP-annotated camera parameters to the Depth Anything metric, providing an metric, stable, and robust scale across training data. By aligning relative-scale camera parameters to metric scales, we can condition the video generation model on accurately scaled camera trajectories, achieving greater control and scene consistency across diverse real-world images.

During inference, RealCam-I2V provides an interactive interface where users can intuitively design camera trajectories by drawing within the reconstructed 3D scene of the reference image. This interface renders preview videos of the trajectory in a static scene, offering users real-time feedback and greater control over camera movement. This interactive feature enhances usability, allowing precise trajectory control even for users without specialized knowledge of scene depth.
To further improve video quality and control precision, we introduce scene-constrained noise initialization as a mechanism to shape the camera movement in its high-noise stages. By using the preview video of the static 3D scene, RealCam-I2V injects scene-visible regions with controlled noise, guiding the video diffusion model’s early generation stages. This high-noise feature constrains the initial layout and camera dynamics, providing a strong foundation for the remaining denoising stages. As denoising progresses, the condition-based approach, trained on metric-scale camera trajectories, preserves global layout and completes the dynamic scene in previously unseen regions. This approach maintains the video diffusion model’s capacity for dynamic content generation while ensuring accurate, coherent camera control.

Our experimental results show that RealCam-I2V achieves significant performance gains in video quality and controllability. When relative scales are aligned to metric scales, models such as MotionCtrl, CameraCtrl, and CamI2V see substantial improvements in video quality. Furthermore, with the introduction of scene-constrained noise initialization, RealCam-I2V surpasses state-of-the-art performance benchmarks, particularly on datasets like RealEstate10K \cite{zhou2018stereo} and out-of-domain images. These results demonstrate the effectiveness of our approach in both controlled and diverse real-world settings.
In summary, our contributions are as follows:
\begin{itemize}
    \item We identify scale inconsistencies and real-world usability challenges in existing trajectory-based methods and introduce a simple yet effective monocular 3D reconstruction into the preprocessing step of the generation pipeline, serving as a reliable intermediary reference for both training and inference.
    \item  With reconstructed 3D scene, we enable metric-scale training and provide an interactive interface during inference to easily design camera trajectories with preview feedback, along with proposed scene-constrained noise shaping to significantly enhance scene consistency and camera controllability.
    \item Our method overcomes critical real-world application challenges and achieves substantial improvements on the RealEstate10K dataset, establishing a new sota benchmark both in video quality and control precision.
\end{itemize}

%% file: sec/2_related_work.tex
\begin{figure*}[!t]
    \centering
    \vspace{-2mm}
    \includegraphics[width=0.98\linewidth]{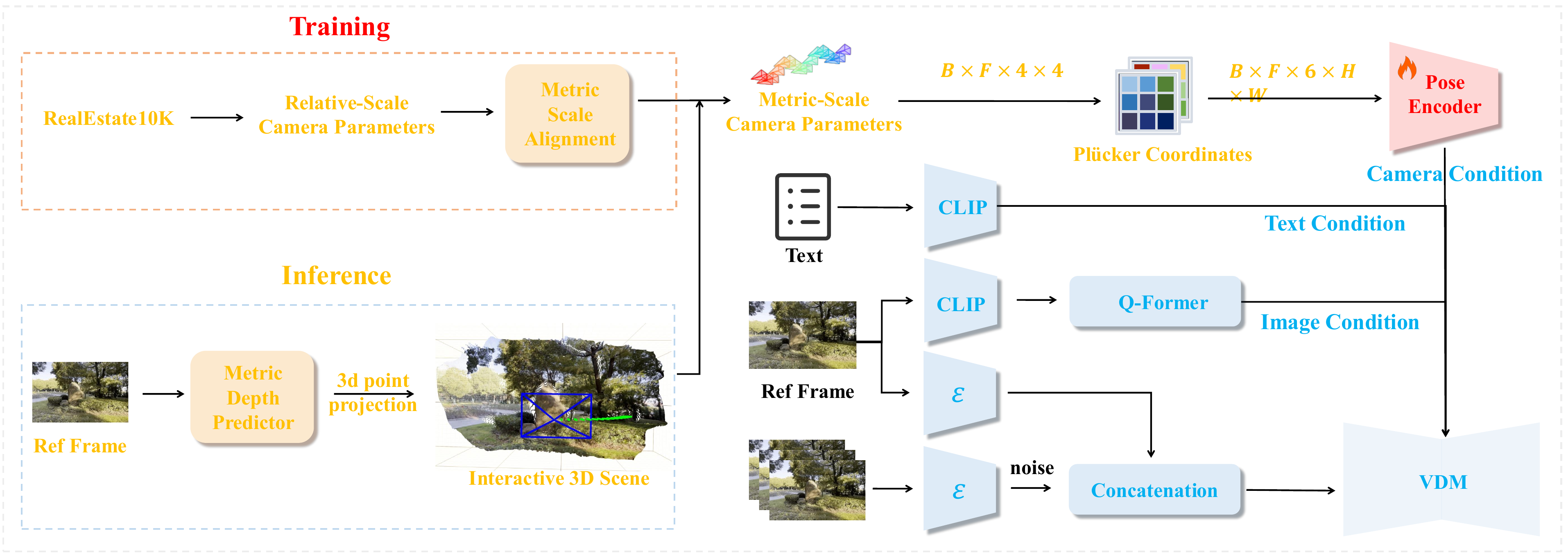}
    \vspace{-5mm}
    \caption{\textbf{RealCam-I2V pipeline.} For training, we align camera parameters from relative scale to metric scale. For inference, we use metric depth estimation to construct the point cloud for users to interactively draw the camera trajectory. Due to the metric scale alignment, the user-given camera trajectory in the 3D scene shares the same scene scale as those in real world.
    }
    \label{fig:pipeline}
    \vspace{-5mm}
\end{figure*}

\section{Related Works}
\label{sec:related_works}

\noindent\textbf{Diffusion-based Video Generation. }
The advancements in diffusion models \cite{rombach2022high, ramesh2022hierarchical, zheng2022entropy} have led to significant progress in video generation, gradually forming two major categories of foundational tasks: text-to-video (T2V) and image-to-video (I2V). 
Due to the lack of high-quality video-text datasets \cite{Blattmann2023, Blattmann2023a}, previous outstanding works such as AnimateDiff \cite{Guo2023}, Align Your Latents \cite{Blattmann2023a}, PYoCo \cite{ge2023preserve}, Emu Video \cite{girdhar2023emu}, LVDM \cite{he2022latent}, VideoCrafter \cite{chen2023videocrafter1, chen2024videocrafter2}, ModelScope \cite{wang2023modelscope}, LAVIE \cite{wang2023lavie}, and VideoFactory \cite{wang2024videofactory} have facilitated text-to-video (T2V) generation by incorporating Motion Blocks into existing text-to-image \cite{podell2023sdxl,rombach2022high,yu2022scaling} models.
Building on these T2V efforts, several works introduced images as control signals, leading to the development of a series of I2V models. Representative examples include SVD \cite{Blattmann2023}, SEINE \cite{chen2023seine}, PixelDance \cite{zeng2024make}, PIA \cite{zhang2024pia}, I2VGen-XL \cite{zhang2023i2vgen}, DynamicCrafter \cite{Xing2023}, and Moonshot \cite{zhang2024moonshot}.
With the improvement in video data \cite{chen2024panda,nan2024openvid,wang2024koala,ju2024miradata} and the rapid development of the DiT architecture \cite{Peebles2023, Ma2024, Yu2024}, recent works such as CogVideoX \cite{yang2024cogvideox}, Sora \cite{videoworldsimulators2024}, HunyuanVideo \cite{kong2024hunyuanvideo}, GoKu \cite{chen2025goku}, Open-Sora \cite{zheng2024open}, Open-Sora-Plan \cite{lin2024open}, Vidu \cite{bao2024vidu}, Lumina-T2X \cite{gao2024lumina}, Vchitect-2.0 \cite{fan2025vchitect}, RepVideo \cite{si2025repvideo} and Step-Video-T2V\cite{ma2025step} have significantly enhanced the quality of video generation. These advancements demonstrate the potential of Video Diffusion Models as realistic world simulators.

\noindent\textbf{Controllable Generation.}
Controllable generation has become a central focus in both image \citep{Zhang2023,jiang2024survey, Mou2024, Zheng2023, peng2024controlnext, ye2023ip, wu2024spherediffusion, song2024moma, wu2024ifadapter} and video \citep{gong2024atomovideo, zhang2024moonshot, guo2025sparsectrl, jiang2024videobooth} generation, enabling users to direct the output through various types of control. A wide range of controllable inputs has been explored, including text descriptions, pose \citep{ma2024follow,wang2023disco,hu2024animate,xu2024magicanimate}, audio \citep{tang2023anytoany,tian2024emo,he2024co}, identity representations \citep{chefer2024still,wang2024customvideo,wu2024customcrafter,wu2024videomaker,zhao20253d,li2024personalvideo}, trajectory \citep{yin2023dragnuwa,chen2024motion,li2024generative,wu2024motionbooth, namekata2024sg}.

\noindent\textbf{Text-based Camera Control.}
Text-based camera control methods use natural language descriptions to guide camera motion in video generation. AnimateDiff \cite{Guo2023} and SVD \cite{Blattmann2023} fine-tune LoRAs \cite{hu2021lora} for specific camera movements based on text input. 
Image conductor\cite{li2024image} proposed to separate different camera and object motions through camera LoRA weight and object LoRA weight to achieve more precise motion control.
In contrast, MotionMaster \cite{hu2024motionmaster} and Peekaboo \cite{jain2024peekaboo} offer training-free approaches for generating coarse-grained camera motions, though with limited precision. VideoComposer \cite{wang2024videocomposer} adjusts pixel-level motion vectors to provide finer control, but challenges remain in achieving precise camera control.

\noindent\textbf{Trajectory-based Camera Control.}
MotionCtrl \cite{Wang2024Motionctrl}, CameraCtrl \cite{He2024Cameractrl}, and Direct-a-Video \cite{yang2024direct} use camera pose as input to enhance control, while CVD \cite{kuang2024collaborative} extends CameraCtrl for multi-view generation, though still limited by motion complexity. To improve geometric consistency, Pose-guided diffusion \cite{tseng2023consistent}, CamCo \cite{Xu2024}, and CamI2V \cite{zheng2024cami2v} apply epipolar constraints for consistent viewpoints. VD3D \cite{bahmani2024vd3d} introduces a ControlNet\cite{Zhang2023}-like conditioning mechanism with spatiotemporal camera embeddings, enabling more precise control.
CamTrol \cite{hou2024training} offers a training-free approach that renders static point clouds into multi-view frames for video generation. Cavia \cite{xu2024cavia} introduces view-integrated attention mechanisms to improve viewpoint and temporal consistency, while I2VControl-Camera~\cite{Feng2024I2VControlCameraPV} refines camera movement by employing point trajectories in the camera coordinate system. 
Recently, 4DiM~\cite{watson2024controlling} and AC3D~\cite{bahmani2024ac3d} also leverage monocular metric depth estimator to tackling the issues of scale inconsistency.
Despite these advancements, challenges in maintaining camera control and scene-scale consistency remain, which our method seeks to address.

%% file: sec/3_method.tex
\section{Method}
\label{sec:method}


Our overall pipeline is shown in~\cref{fig:pipeline}. For training, we align camera parameters from relative scale to metric scale. For inference, we use metric depth estimation to construct the point cloud for users to interactively draw the camera trajectory. Due to the metric scale alignment, the user-given camera trajectory in the 3D scene shares the same scene scale as those in real world.


\subsection{Camera-controlled Image-to-Video Model} 

Instead of directly modeling the video \( x \), the latent representation \( z = \mathcal{E}(x) \) is used for training. The diffusion model \( \epsilon_\theta \) learns to estimate the noise \( \epsilon \) added at each timestep \( t \), conditioned on both a text prompt \( c_\text{txt} \), a reference image \( c_\text{img} \), and camera condition \( c_\text{cam} \), with \( t \in \mathcal{U}(0, 1) \). The training objective simplifies to a reconstruction loss defined as:
\begin{equation}
    \mathcal{L} = \mathbb{E}_{z, c_\text{txt}, c_\text{img}, c_\text{cam}, \epsilon, t} 
    \left[ \left\| \epsilon - \epsilon_\theta\left(z_t, c_\text{txt}, c_\text{img}, c_\text{cam}, t\right) \right\|_2^2 \right],
\label{eq:diffusion_loss}
\end{equation}
where \( z \in \mathbb{R}^{F \times H \times W \times C} \) represents the latent code of a video, with \( F, H, W, C \) corresponding to frame count, height, width, and channel dimensions. The noise-corrupted latent code \( z_t \), derived from the ground-truth latent \( z_0 \), is expressed as:
\begin{equation}
    z_t = \alpha_t z_0 + \sigma_t \epsilon,
\end{equation}
where \( \sigma_t = \sqrt{1 - \alpha_t^2} \). Here, \( \alpha_t \) and \( \sigma_t \) are hyperparameters governing the diffusion process.



\begin{figure}[!t]
    \centering
    \vspace{-1mm}
    \includegraphics[width=1.0\linewidth]{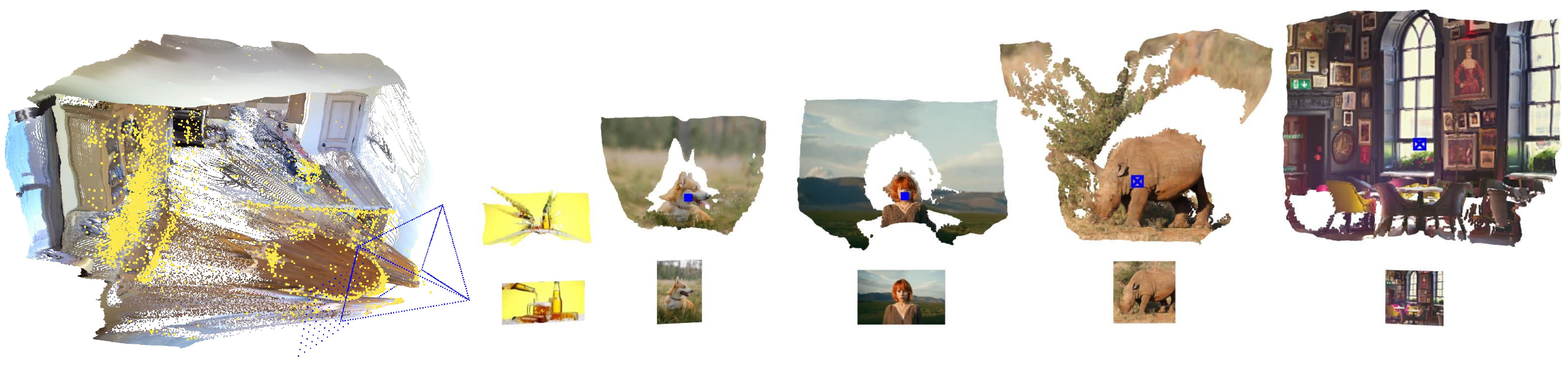}
    \vspace{-6mm}
    \caption{
    \textbf{Scene scale mismatch.}
    Point clouds reconstructed from metric depth estimation (RGB) are robust and unified, whereas SfM reconstructions (yellow) are relative-scale that may vary across frames.
    Our alignment enables relative-to-metric conversion of scene scale for real-world applications.
    }
    \label{fig:align}
    \vspace{-2mm}
\end{figure}

\begin{figure}[!t]
    \centering
    \includegraphics[width=1.0\linewidth]{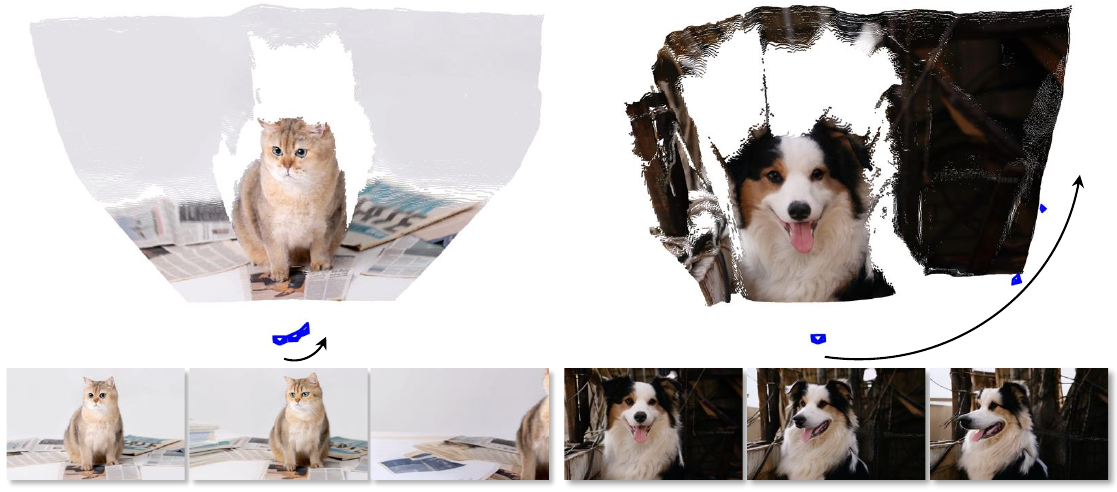}
    \vspace{-6mm}
    \caption{
        \textbf{Camera trajectory ambiguity.}
        The relative scene-scale measurement fundamentally hinders models from learning physically consistent camera motion.
    } 
    \label{fig:scale_ambiguity}
    \vspace{-5mm}
\end{figure}

\subsection{Metric Scene-scale Alignment}


\noindent\textbf{Metric Depth Estimation.}
\label{sec:3d_reconstruction}
To obtain a depth map from a given input image, we use a metric depth predictor \( f_{\text{depth}} \), which takes the RGB image \( I \) as input and outputs the corresponding depth map \( D(u, v) \). 
The prediction process is formulated as:
\[
D(u, v) = f_{\text{depth}}(I),
\]
where $I$ is the input RGB image and $ D(u, v) $ is the predicted depth value for each pixel at coordinates $ (u, v) $. This predicted depth map \( D(u, v) \) serves as the foundation for projecting the image into 3D space, allowing us to construct a point cloud in the camera coordinate system.
The camera intrinsics matrix \( K \) is defined as:
\[
K = 
\begin{bmatrix}
f_x & 0 & c_x \\
0 & f_y & c_y \\
0 & 0 & 1
\end{bmatrix},
\]
where \( f_x \) and \( f_y \) are the focal lengths along the \( x \) and \( y \) axes,
\( (c_x, c_y) \) is the principal point of the camera.

\noindent\textbf{Metric Scene-Scale Alignment.}
To convert camera extrinsics from world-to-camera to an metric-scale camera-to-world representation, we defines that the world-to-camera extrinsics matrix \( F_{\text{w2c}} \in \mathbb{R}^{4 \times 4} \) is inverted to obtain the corresponding camera-to-world matrix
\(
F_{\text{c2w}} = F_{\text{w2c}}^{-1}.
\)
To express the transformations relative to the first frame, each \( F_{\text{c2w}} \) is left-multiplied by the camera-to-world matrix of the inverse of first frame \( F_{\text{c2w, 1}} \):
\[
c_{\text{cam}} = F_{\text{c2w,1}}^{-1} \cdot F_{\text{c2w}}.
\]
Here, \( c_{\text{cam}} \in \mathbb{R}^{F \times 4 \times 4} \) represents the camera-to-world transformations aligned relative to the first frame. However, the translation component of \( c_{\text{cam}} \) remains in a relative scene scale.
To convert the relative translation to an metric scale, we align the metric 3D point cloud reconstructed by Depth Anything (metric version) with the 3D point cloud reconstructed by COLMAP (Structure-from-Motion), as shown in~\cref{fig:align} and~\cref{fig:scale_ambiguity}. The alignment process yields a scale factor \( \alpha \) and is applied to the translation component of \( c_{\text{cam}} \), resulting in an metric-scale camera-to-world transformation:
\[
c_{\text{cam}}^{\text{metric}} = 
\begin{bmatrix}
R & \alpha \cdot T \\
0 & 1
\end{bmatrix},
\]
where \( R \) is the rotation matrix, \( T \) is the relative translation vector.
The resulting \( c_{\text{cam}}^{\text{metric}} \in \mathbb{R}^{F \times 4 \times 4} \) represents the camera-to-world transformations with metric scene scale, enabling robust and accurate real-world applications.

\begin{figure}[!t]
    \centering
    \vspace{-1mm}
    \includegraphics[width=1.0\linewidth]{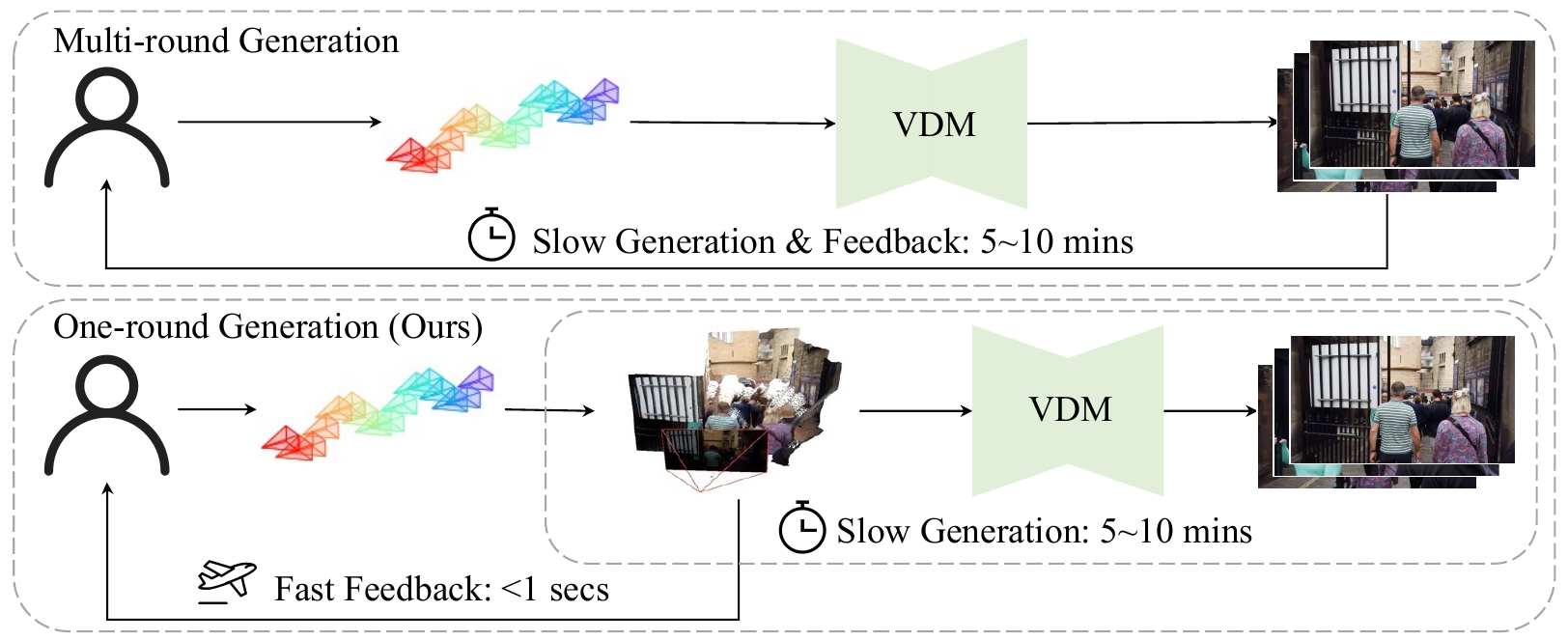}
    \vspace{-6mm}
    \caption{
        \textbf{One-round generation versus multi-round generation.}
        Our framework decouples camera adjustment via interactive 3D scenes, enabling fast feedback before slow generation.
    }
    \label{fig:one_round}
    \vspace{-5mm}
\end{figure}

\subsection{One-round Generation in Inference}
Users are allowed to draw camera trajectory in the constructed 3D scene and can easily preview the rendering video of expected camera trajectory, free from the costly generation via video diffusion models, as shown in~\cref{fig:one_round}.


\noindent\textbf{3D Point Projection For Interaction.}
Given a depth map \( D(u, v) \), 
the projected 3D coordinates in the camera coordinate system, \( \mathbf{p}_c = (x_c, y_c, z_c)^T \), are computed as:
\[
\mathbf{p}_c = {\bf D}(u, v) \cdot K^{-1} \cdot 
\begin{bmatrix} 
u \\ 
v \\ 
1 
\end{bmatrix}.
\]
Here \( \begin{pmatrix} u, v, 1 \end{pmatrix} \) represents the homogeneous coordinates of the pixel,
  \( K^{-1} \) is the inverse of the intrinsic matrix, which maps pixel coordinates to normalized image coordinates.
By applying this transformation to all pixels in the depth map, we obtain a set of 3D points \( \{ \mathbf{p}_c \} \) in the camera coordinate system. 

\begin{figure}[!t]
    \centering
    \vspace{-2mm}
    \includegraphics[width=1.0\linewidth]{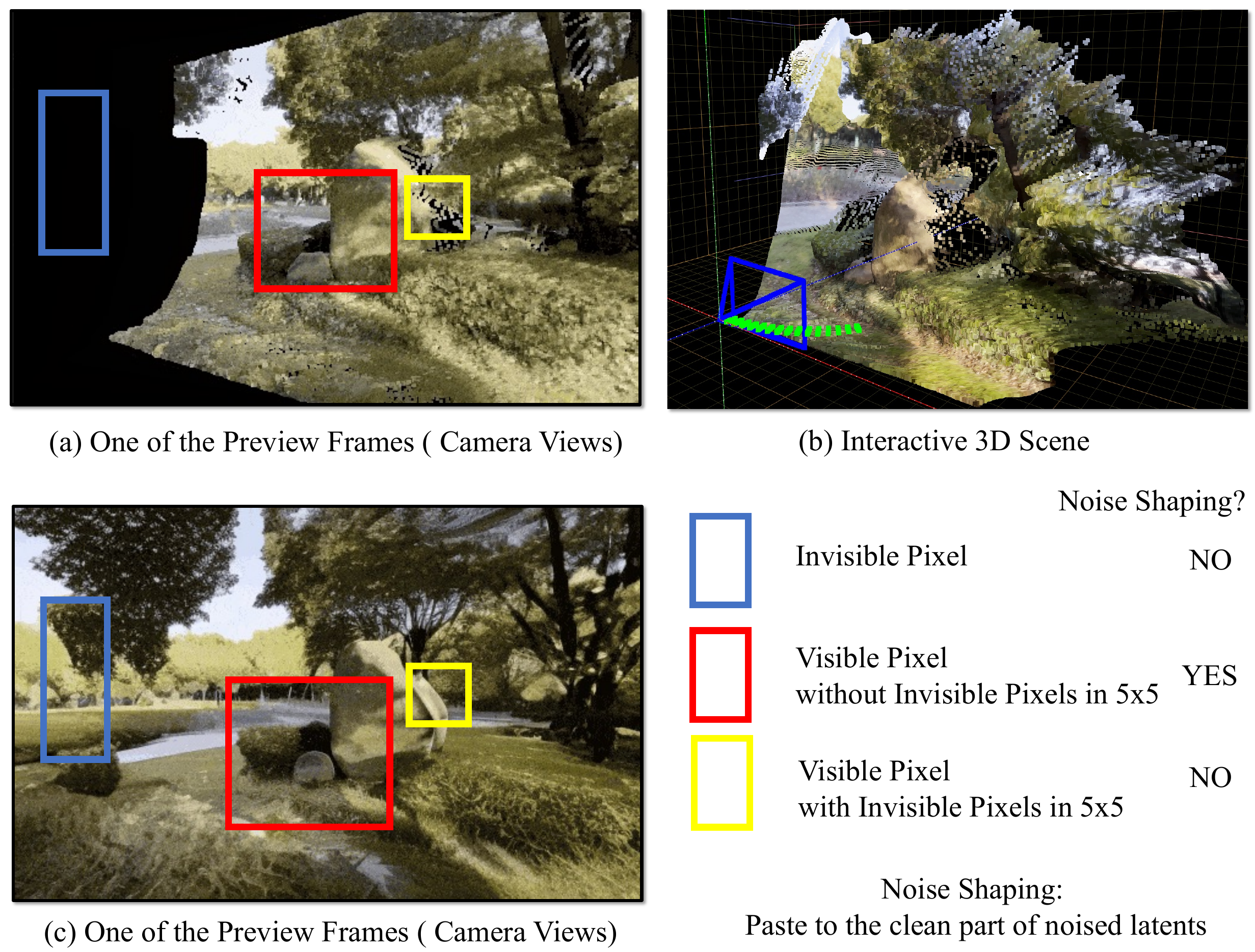}
    \vspace{-6mm}
    \caption{
    \textbf{Pixel selection for noise shaping.}
    Selected pixels are pasted onto the clean part (predicted $z_0$) of noised latent $z_t$.
    Empirically, applying noise shaping at high noise levels $t>0.9$ achieves the trade-off of camera control and dynamics.
    }
    \label{fig:noise_shaping}
    \vspace{-6mm}
\end{figure}

\begin{figure}[!t]
    \centering
    \vspace{-1mm}
    \includegraphics[width=1.0\linewidth]{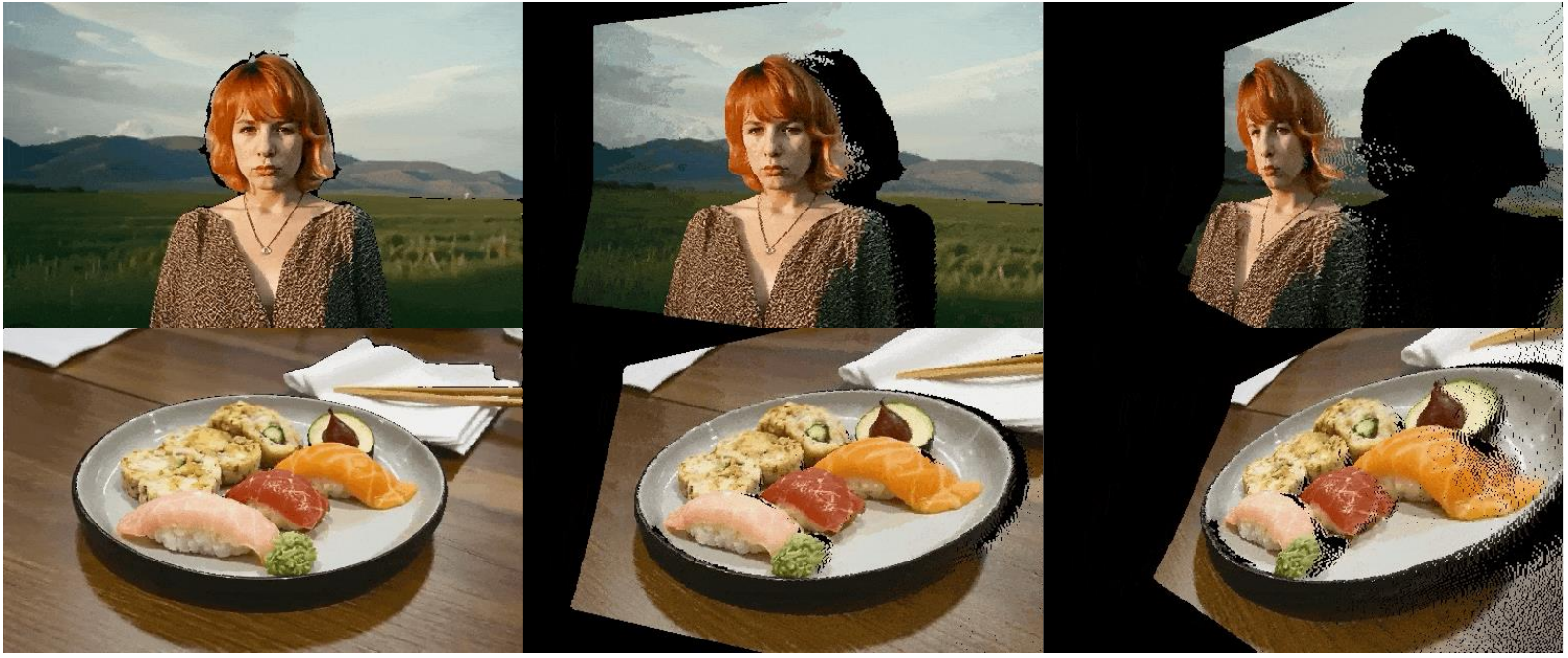}
    \vspace{-6mm}
    \caption{
    \textbf{Preview video rendering for user-expected camera trajectory.}
    It also serves as the reference video in our proposed scene-constrained noise shaping.
    }
    \label{fig:rendering}
    \vspace{-2mm}
\end{figure}

\noindent\textbf{Preview Video Rendering for Camera Motion.}
With constructed 3D point cloud, we can render a preview video of camera traces. As shown in~\cref{fig:rendering}, the preview video functions as a reference video for validation.

\noindent\textbf{Scene-constrained Noise Shaping.}
Inspired by previous works SDEdit~\cite{meng2021sdedit} and DDIM inversion~\cite{song2020denoising}, noised features $z_t$ can be used for shaping the  layout, camera control of the entire image, especially at timestep with high-level noise.
We propose \textit{scene-constrained noise shaping}, which utilizes preview videos generated along user-defined trajectories in the interactive 3D scene. Each frame of the preview video is treated as a reference frame and provided to the generation process during the high-noise stage. The reference frame's pixels are overlaid onto the model-predicted \( z_0 \) to achieve the shaping effect.

Next, we detail the process for selecting the pixels to be referenced. As illustrated in~\cref{fig:noise_shaping}, the primary criterion is that a pixel must be visible under the current camera viewpoint in the preview video. To mitigate issues such as holes caused by inaccurate depth predictions, we apply an additional filtering rule: if a visible pixel's \( k \times k \) neighborhood contains any invisible pixels, it is considered to lie on an object's edge and potentially affected by depth prediction errors. Such pixels are excluded from selection.
Finally, we define the noise shaping process as the formula:
\[
z_t = m \cdot (\alpha_{t} z_\text{preview} + \sigma_{t} \epsilon) + (1 - m) \cdot z_t,
\]
where $m$ identifies the selected reference pixels, \( z_\text{preview} \) represents the clean latent features from the preview video. It should be noted that resampling $\epsilon$ in each timestep is important because the motivation of noise shaping is to utilize the clean feature of preview feature to shape the low-frequency information of the generation process while the random sampled $\epsilon$ may cover some of the valid information in each timestep if it is sampled to be a large value.

\begin{figure}[!t]
    \centering
    \vspace{-1mm}
    \includegraphics[width=1.0\linewidth]{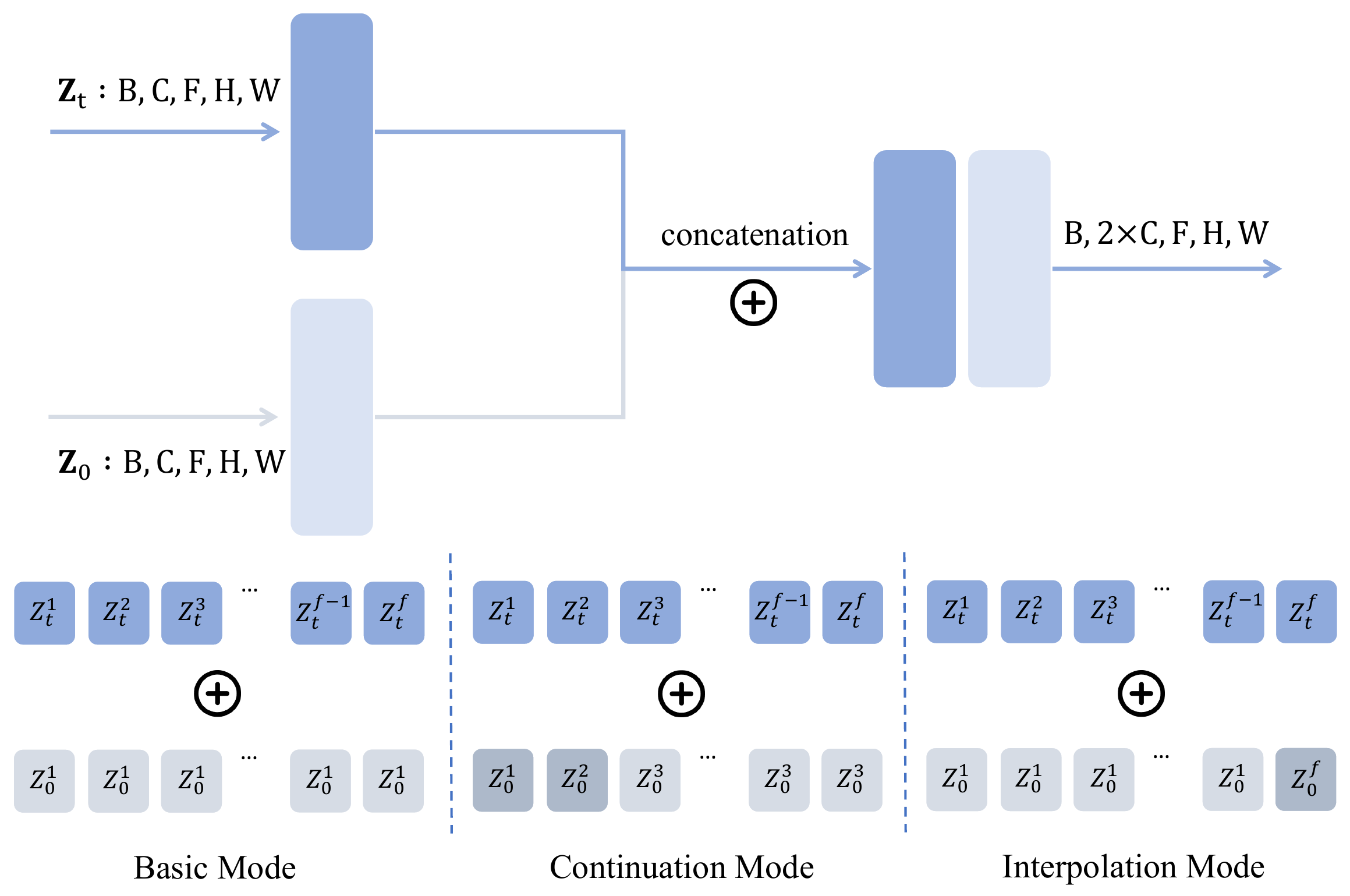}
    \vspace{-6mm}
    \caption{
    \textbf{Concatenation strategies for different tasks.}
    Basic mode, interpolation mode, and continuation mode can be supported with only minor modification.
    }
    \label{fig:concatenation_mode}
    \vspace{-5mm}
\end{figure}

\input{tables/quantative_comparision_with_sota.tex}

\subsection{Interpolation, Loop and Continuation}

To support different tasks, including interpolation, looping, and continuation for long video generation, we train video diffusion model with different input concatenation mode, as shown in~\cref{fig:concatenation_mode}.
Given a video latents \( z \in \mathbb{R}^{F \times H \times W \times C} \), we define the noised latents of $f$-th frame at timestep $t$ as $z_{t}^{f}$. We then select $i$-th clean frame as the condition frame $z^{i}_0$. For interpolation mode, we define $z^{f-1}_0$ as the end condition frame. For continuation mode, we define all $1~i$-th as condition frame.

%% file: tables/quantative_comparision_with_sota.tex
\begin{table*}[htbp]
    \centering
    \newcommand{\gray}[1]{\textcolor{black!40}{#1}}
    \resizebox{\linewidth}{!}{
    \begin{tabular}{l|c|cc|cc|cc}
        \toprule
        \multirow{2}{*}{Method} & \multirow{2}{*}{RotErr~$\downarrow$} & \multicolumn{2}{c|}{TransErr~$\downarrow$} & \multicolumn{2}{c|}{CamMC~$\downarrow$} & \multicolumn{2}{c}{FVD~$\downarrow$} \\
        & & Relative Scale & Metric Scale & Relative Scale & Metric Scale & VideoGPT & StyleGAN \\

        \Xhline{0.5pt}
        \cmidrule{1-8}

        DynamiCrafter~\cite{Xing2023}           & 3.3415 & 9.8024 & 14.135 & 11.625 & 15.726 & 106.02 & 92.196 \\
        MotionCtrl$^*$~\cite{Wang2024Motionctrl}  & 1.0527 & 2.2860 & 6.8182 & 2.9312 & 7.2272 & 70.292 & 60.845 \\
        CameraCtrl$^*$~\cite{He2024Cameractrl}   & 0.7373 & 1.7619 & 5.5090 & 2.1644 & 5.7648 & 69.202 & 58.900 \\
        CamI2V$^*$~\cite{zheng2024cami2v}        & 0.4120 & 1.3409 & 3.2934 & 1.5291 &3.4230  & 62.439 & 53.361 \\
        
        \cmidrule{1-8}
        
        \rowcolor{gray!15}
        RealCam-I2V (Ours)                    & \textbf{0.3884} & \textbf{1.2943} & \textbf{2.2317} & \textbf{1.4628} & \textbf{2.3609} & \textbf{53.718} & \textbf{45.460} \\
        \textit{w.r.t.} CamI2V                                        & \red{+5.73\%} & \red{+3.48\%} & \red{+32.24\%} & \red{+4.33\%} & \red{+31.03\%} & \red{+13.97\%} & \red{+14.81\%}\\

        \bottomrule
    \end{tabular}
    }
    \vspace{-2mm}
    \caption{
        \textbf{Quantitative comparison with SOTA methods.} 
        Our approach excels all baselines on both relative and metric results, while coherently improve visual quality of generated videos.
        We observe over 30\% improvement on metric scale results and over 10\% improvement on FVD.
        * denotes our reproduced results on DynamiCrafter~\cite{Xing2023}. 
        \textbf{Best} and \underline{second best} results are highlighted respectively. 
    }
    \label{tab:comparison}
    \vspace{-5mm}
\end{table*}

%% file: sec/4_experiment.tex
\section{Experiments}
\label{sec:experiments}

\subsection{Setup}

\noindent\textbf{Dataset.} We train our model on RealEstate10K~\cite{zhou2018stereo}, which contains 70,000 video clips with well-annotated relative-scale camera poses. 
For metric scene scale alignment, we run COLMAP~\cite{Schonberger2016} point triangulator on each video clip with fixed camera intrinsics and extrinsics directly from RealEstate10K, obtaining the sparse point cloud of the reconstructed scene.
For metric depth predictor, we choose Depth Anything V2 \cite{depth_anything_v2} Large Indoor, which is fine-tuned on metric depth estimation.
We then calculate per-point relative-to-metric scaling factor against metric depth prediction.
We term the median value of scaling factors in a frame as the frame-level factor for robustness. 
To make stable training, we eliminate outliers of video clips whose maximum frame-level scaling factors are among the first 2\% with too small values or the last 2\% with too large values, assuming sorted in ascending order.
The same quantile filtering strategy is applied on the minimum frame-level scaling factors of video clips.
It remains 58,000 video clips for training and another 6,000 for testing.
To further demonstrate the robustness of the proposed method while enhancing dynamics and diveristy, we reproduce RealCam-I2V on our open-sourced RealCam-Vid~\cite{zheng2025realcam}, a high-resolution video dataset with aligned metric-scale camera annotations and diverse scene dynamics. More experimental details can be found in ~\cref{app:robustness}.

\noindent\textbf{Implementation Details.} We choose DynamiCrafter~\cite{Xing2023} as our image-to-video (I2V) base model and seamlessly integrate proposed RealCam-I2V into it as a plugin. 
During depth-aligned training, we freeze all parameters of the base model and the depth predictor, while only parameters of proposed method are trainable. 
We follow DynamiCrafter to sample 16 frames from each single video clip while perform resizing, keeping the aspect ratio, and center cropping to fit in our training scheme.
We train the model with a random frame stride ranging from 1 to 10 and take random condition frame as data augmentation.
We fix the frame stride to 8 and always use the first frame as the condition frame for inference.
We supervise \(\epsilon\)-prediction on the model of 256\(\times\)256 resolution and \(v\)-prediction on the model of 512\(\times\)320 resolution respectively, following the pre-training scheme of DynamiCrafter.
We apply the Adam optimizer with a constant learning rate of \(1 \times 10^{-4}\) with mixed-precision fp16 and ZeRO-1.

\input{tables/ablation_study}
\input{tables/main_Vbench-I2V}

\subsection{Metrics}

We follow previous works~\cite{Wang2024Motionctrl, He2024Cameractrl, xu2024camco, zheng2024cami2v} to evaluate camera-controllablity by RotErr, TransErr and CamMC on their estimated camera poses using structure-from-motion (SfM) methods, \eg COLMAP~\cite{Schonberger2016} and GLOMAP~\cite{pan2024glomap}.
We convert the camera pose of each frame in a video clip to be relative to the first frame as canonicalization.
We denote the \(i\)-th frame relative camera-to-world matrix of ground truth as \(\{R^{3\times3}_i,T^{3\times1}_i\}\), and that of generated video as \(\{\tilde{R}^{3\times3}_i,\tilde{T}^{3\times1}_i\}\).
We randomly select 1,000 samples from test set for evaluation.
We sum up per-frame errors as the scene-level result for camera metrics.
Inspired by ~\citet{zheng2024cami2v}, we repetitively conduct 5 individual trials on each video clips for camera-control metrics to reduce the randomness introduced by SfM tools.
Metrics of one video clip are averaged on successful trials at first for later sample-wise average to get final results.

\noindent\textbf{RotError~\textnormal{\cite{He2024Cameractrl, xu2024camco, zheng2024cami2v}}.} We calculate camera rotation errors by the relative angle between generated videos and ground truths in radians for rotation accuracy. 
\begin{equation}
    {\rm RotErr} = \sum_{i=1}^n \arccos{\frac{\mathop{\rm tr}(\tilde{R}_i R_i^{\rm T}) - 1}{2}}    
\end{equation}

\noindent\textbf{TransError~\textnormal{\cite{He2024Cameractrl, xu2024camco, zheng2024cami2v}}.} For relative TransErr, we perform scene scale normalization on the camera positions of each video clip.
The scene scale of generated video \(\tilde{s}_i\) and grouth truth \(s_i\) are individually calculated as the \({\cal L}_2\) distance from the first camera to the farthest one for each video clip.
For absolute TransErr, we normalize both the video clip to the scene scale of ground truth video, \ie \(\tilde{s}_i=s_i\).
\begin{equation}
    {\rm TransErr} = \sum_{i=1}^n{\left\Vert \frac{\tilde{T}_{i}}{\tilde{s}_i} - \frac{T_{i}}{s_i} \right\Vert_2}
\end{equation}

\noindent\textbf{CamMC~\textnormal{\cite{Wang2024Motionctrl}}.} We perform the same scene scale normalization for relative and metric results as TransError, and evaluate the overall camera pose accuracy by directly calculating \({\cal L}_2\) similarity on camera-to-world matrices.
\begin{equation}
    {\rm CamMC} = \sum_{i=1}^n{\left\Vert \begin{bmatrix} \tilde{R}_i \,\Big|\, \dfrac{\tilde{T}_{i}}{\tilde{s}_i} \end{bmatrix}^{3\times4} - \begin{bmatrix} R_i \,\Big|\, \dfrac{T_{i}}{s_i} \end{bmatrix}^{3\times4} \right\Vert_2}
\end{equation}


\noindent\textbf{FVD~\textnormal{\cite{Unterthiner2018}}.} We also assess the visual quality of generative videos by the distribution distance between generated videos and ground truths.

\subsection{Comparison with SOTA Methods}

We compare RealCam-I2V with models that either lack camera-condition training (DynamiCrafter~\cite{Xing2023}) or incorporate camera-condition training, namely MotionCtrl~\cite{Wang2024Motionctrl} (3\(\times\)4 camera extrinsics), CameraCtrl~\cite{He2024Cameractrl} (Plücker embedding constructed from camera extrinsics and intrinsics as side input), and CamI2V~\cite{zheng2024cami2v} (current SOTA using Plücker embedding and epipolar attention between all frames), as shown in~\cref{tab:comparison}.
We report all results using DynamiCrafter as the base model.

Our method demonstrates significant improvements in visual quality (FVD) and camera control metrics (TransErr, RotErr, CamMC), with particularly over 30\% gains on metric scale results.
However, these improvements are not fully captured by the RealEstate10K dataset, which contains mostly static scenes.

\begin{figure*}[!ht]
    \centering
    \vspace{-2mm}
    \includegraphics[width=1.0\linewidth]{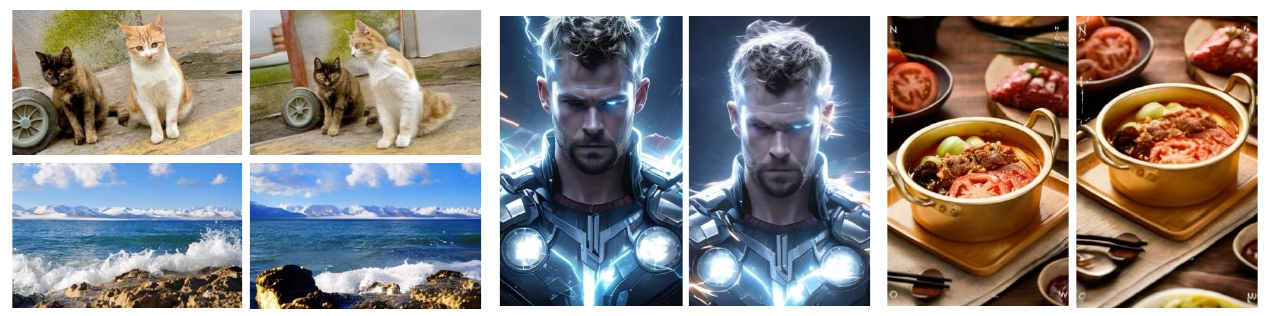}
    \vspace{-7mm}
    \caption{
        \textbf{Visualization on various domains in real life scenarios.}
        Despite training on RealEstate10K~\cite{zhou2018stereo}, our method can generalize naturally to out-of-domain images, including pets, landscape, anime, food and etc.
    } 
    \label{fig:various_domain}
    \vspace{-4mm}
\end{figure*}

\begin{figure}[!t]
    \vspace{-1mm}
    \centering
    \includegraphics[width=1.0\linewidth]{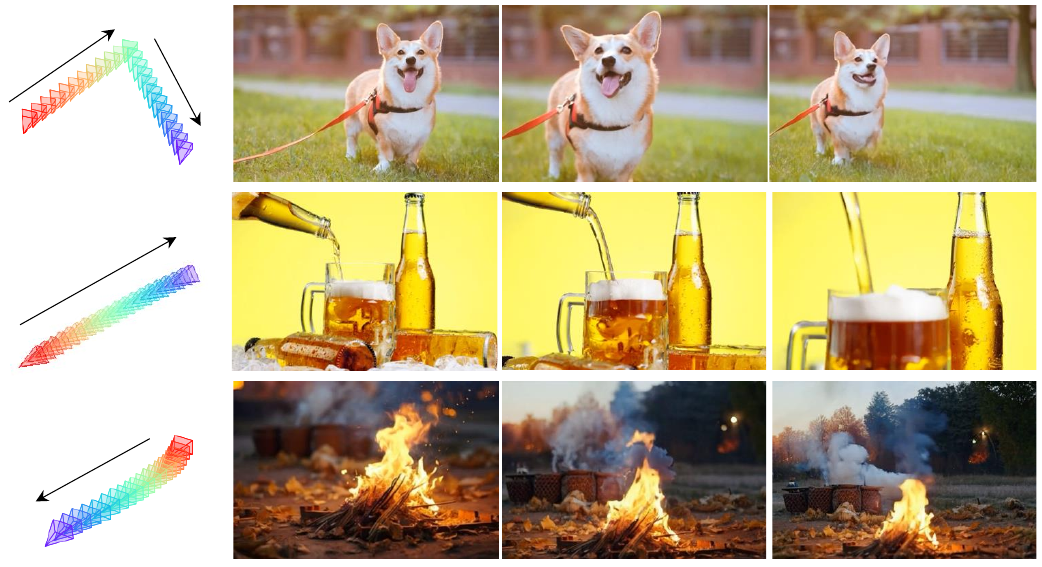}
    \vspace{-6mm}
    \caption{
        \textbf{Visualization on complex trajectory, large movement and video dynamism.}
        Our framework achieves precise trajectory adherence for complex camera motion paths while preserving high-fidelity video generation with dynamics.
    }
    \vspace{-5mm}
    \label{fig:complex_trajectory}
\end{figure}

\subsection{Ablation Study}


To validate our proposed method, we conducted a comprehensive ablation study on Metric Scene-scale Alignment (MSA) and Scene-constrained Noise Shaping (SNS).

\noindent\textbf{Effect of MSA only.}
As shown in~\cref{tab:ablation}, compared to relative scale training, metric scale training mitigates scale ambiguity in~\cref{fig:scale_ambiguity} and yields notable improvements, especially on metric scale results. 
It implies that models trained on metric-scale data can more accurately capture true-to-scale translations and better understand camera rotations within a realistic spatial framework.
The metric scene scale enhances robustness and compatibility, ensuring that the framework adapts effectively to real-world input and applications.
This approach allows for interaction within a unified scale, enabling intuitive user control over camera actions. 


\noindent\textbf{Effect of MSA together with SNS.}
Adding scene-constrained noise shaping to a model with metric scale training yields substantial gains in video quality and camera controllability.
This improvement is evident in both camera metrics and FVD. 
The synergy of metric-scale training and noise shaping ensures robust and precise control in diverse scenarios.
This combined approach delivers noticeably better dynamics compared to using noise shaping alone. 
Large camera movements, rotations, and rapid transitions, which previously struggled to maintain consistency and realism, now work seamlessly. 
This improvement underscores the strength of integrating metric scale training with noise shaping for complex motion scenarios.






\noindent\textbf{Effect of SNS only.}
As shown in~\cref{tab:ablation}, scene-constrained noise shaping can be used as the only method for camera control when applied to a base model not trained with any camera conditions.
It provides nearly 50\% reduction on DynamiCrafter. 
However, this method performs poorly compared to the combined method with metric scale training.
It also introduces challenges in parameter selection:
Applying noise shaping only in the high noise stages limits camera control in lower noise stages, while extending noise shaping to mid-noise phase can suppress video dynamics, resulting in static video output.
This limitation affects the fluidity and responsiveness of camera movements, making our combined approach preferable for applications requiring natural dynamics.






\subsection{Application}




As illustrated in \cref{fig:various_domain} and \cref{fig:complex_trajectory}, we demonstrate the versatility of our method through visualization results across various applications. 
Additionally, our results include camera-controlled loop video generation, generative frame interpolation, and smooth scene transitions, highlighting the robustness of our approach. 
These visualizations showcase two major breakthroughs: first, our method achieves a real-world application breakthrough by addressing challenges like training-inference scale inconsistency and low usability, ensuring improved robustness and compatibility with real-world images. 
Second, our framework exhibits superior performance in complex camera motions, handling large and rapid movements, rotations, and dynamics more effectively than existing methods. 

%% file: tables/ablation_study.tex
\begin{table*}[!ht]
    \centering
    \newcommand{\gray}[1]{\textcolor{black!45}{#1}}
    \resizebox{\linewidth}{!}{
    \begin{tabular}{l|cc|c|cc|cc|cc}
        \toprule
        \multirow{2}{*}{Method} & \multirow{2}{*}{MSA} & \multirow{2}{*}{SNS} & \multirow{2}{*}{RotErr~$\downarrow$} & \multicolumn{2}{c|}{TransErr~$\downarrow$} & \multicolumn{2}{c|}{CamMC~$\downarrow$} & \multicolumn{2}{c}{FVD~$\downarrow$} \\
        &  &  & & Relative Scale & Metric Scale & Relative Scale & Metric Scale & VideoGPT & StyleGAN \\
        
        \Xhline{0.5pt}
        \cmidrule{1-10}

        \multirow{2}{*}{DynamiCrafter~\cite{Xing2023}} 
        &  &            & 3.3415 & 9.8024 & 14.135 & 11.625 & 15.726 & 106.02 & 92.196 \\
        &  & \checkmark & 1.5163 & 6.6392 & 8.4607 & 7.2108 & 8.9505 & 71.942 & 65.014 \\
        
        \cmidrule{1-10}
        
        \multirow{3}{*}{MotionCtrl$^*$~\cite{Wang2024Motionctrl}} 
        &            &            & 1.0527 & 2.2860 & 6.8182 & 2.9312 & 7.2272 & 70.292 & 60.845 \\
        & \checkmark &            & 0.8655 & 2.3342 & 4.2218 & 2.8083 & 4.5984 & 67.130 & 58.311 \\
        & \checkmark & \checkmark & 0.6373 & 2.0725 & 3.2308 & 2.3771 & 3.4721 & 58.885 & 50.111 \\
         
        \cmidrule{1-10}
        
        \multirow{3}{*}{CameraCtrl$^*$~\cite{He2024Cameractrl}} 
        &            &            & 0.7373 & 1.7619 & 5.5090 & 2.1644 & 5.7648 & 69.202 & 58.900 \\
        & \checkmark &            & 0.7042 & 1.9477 & 3.8218 & 2.3007 & 4.0829 & 60.314 & 51.918 \\
        & \checkmark & \checkmark & 0.5436 & 1.7954 & 3.1845 & 2.0336 & 3.3620 & \underline{55.004} & \underline{46.702} \\
         
        \cmidrule{1-10}
        

        
        \multirow{2}{*}{CamI2V$^*$~\cite{zheng2024cami2v}} 
        &            &            & \underline{0.4120} & \underline{1.3409} & 3.2934 & \underline{1.5291} &3.4230  & 62.439 & 53.361 \\
        & \checkmark &            & 0.4243 & 1.3596 & \underline{2.6524} & 1.5539 & \underline{2.8008} & 60.516 & 51.581 \\
        \rowcolor{gray!15}
        RealCam-I2V (Ours) & \checkmark & \checkmark & \textbf{0.3884} & \textbf{1.2943} & \textbf{2.2317} & \textbf{1.4628} & \textbf{2.3609} & \textbf{53.718} & \textbf{45.460} \\

        \bottomrule
    \end{tabular}
    }
    \vspace{-2mm}
    \caption{
        \textbf{Ablation study of RealCam-I2V plugins.} 
        Metric Scene-scale Alignment (MSA) mitigates scale inconsistency for real-world applications, indicating a more stable and unified camera control.
        Scene-constrained Noise Shaping (SNS) solely provides substantial improvements on the base model but is less effective than the combined approach (ours).
        * denotes our reproduced results on DynamiCrafter~\cite{Xing2023}.
        \textbf{Best} and \underline{second best} results are highlighted respectively.
    }
    \label{tab:ablation}
    \vspace{-2mm}
\end{table*}

%% file: tables/main_Vbench-I2V.tex
\begin{table*}[ht!]
  \centering
    \resizebox{\linewidth}{!}{
        \renewcommand{\arraystretch}{1.0} 
        \begin{tabular}{l|c|ccc|cccccc}
            \toprule
            \multirow{2}{*}{Method} 
              & Total  & I2V & I2V  & Camera  & Subject & Background & Motion & Dynamic & Aesthetic  & Imaging\\
              & Score  &  Subject  & Background  & Motion & Consistency & Consistency & Smoothness & Degree & Quality  & Quality  \\
            \Xhline{0.5pt}
            \midrule

            DynamiCrafter~\cite{Xing2023}               & 84.22 & \underline{95.93} & \textbf{95.43} & 22.67 & \textbf{95.44} & \textbf{98.54} & \textbf{98.08} & 34.15 & \underline{58.98} & 62.35  \\

            \midrule
            
            \rowcolor{gray!15}
            RealCam-I2V (Ours)             & \textbf{85.71} & \textbf{96.14} & \underline{95.27} & \textbf{93.32} & \underline{93.96} & \underline{97.58} & \underline{97.66} & 35.77 & \textbf{59.79} & \textbf{63.08}  \\
            \textit{w.o.} SNS              & \underline{84.37} & 94.73 & 93.44 & 87.42 & 90.99 & 96.23 & 97.36 & \textbf{46.75} & 58.37 & 62.91  \\
            \textit{w.o.} MSA             & 83.77 & 94.71 & 92.65 & \underline{91.48} & 91.24 & 96.06 & 97.35 & \underline{36.99} & 58.32 & \underline{63.03}  \\

            \bottomrule
      \end{tabular}
    }
  \vspace{-2mm}
  \caption{
  \textbf{Ablation study on Vbench-I2V~\cite{huang2024vbench++}},  
  investigating how camera-conditioned fine-tuning exclusively on static RealEstate10K~\textnormal{\cite{zhou2018stereo}} data affects the generation quality and camera motion of the base model.
  Notably, even fine-tuned on a static dataset, it preserves dynamics (Dynamic Degree) without compromise.
  Introducing dynamic datasets will better enhance dynamics, we leave it for future work.
  }
  \label{tab:main_Vbench-I2V}
  \vspace{-5mm}
\end{table*}

%% file: sec/5_conclusion.tex
\section{Limitation Analysis and Future Work}
Despite training in static RealEstate10K, as discussed in~\cref{fig:various_domain} and~\cref{fig:complex_trajectory}, our method demonstrate well generalization to videos with dynamic scenes, various styles and domains due to the designed algorithm to best preserve the knowledge of foundation model,  However, the current model still suffers from real-world application mainly due to the limitation of data and there's no suitable dataset designed for camera-controlled video generation, which requires both dynamic scenes and camera movement.

\section{Conclusion}
\label{sec:conclusion}
In this paper, we address the scale inconsistencies and real-world usability challenges in existing trajectory-based camera-controlled image-to-video generation methods. We introduce a simple yet effective monocular 3D reconstruction into the preprocessing step of the generation pipeline, serving as a reliable intermediary reference for both training and inference. With reconstructed 3D scene, we enable absolute-scale training and provide an interactive interface during inference to easily design camera trajectories with preview feedback, along with proposed scene-constrained noise shaping to significantly enhance scene consistency and camera controllability. Our method overcomes critical real-world application challenges and achieves substantial improvements on the RealEstate10K dataset, establishing a new sota both in video quality and control precision.

%% file: sec/supplementary.tex
\clearpage
\setcounter{page}{1}

\def\maketitleappendix
   {
   \newpage
       \twocolumn[
        \centering
        \Large
        \textbf{\thetitle}\\
        \vspace{0.5em}Appendix\\
        \vspace{1.0em}
       ] 
   }
   
\maketitleappendix
\appendix


\input{tables/rebuttal_noise_shaping}
\input{tables/rebuttal_realcam-vid}

\section{Dataset}

The camera trajectory of each video clip from RealEstate10K~\cite{zhou2018stereo} is first derived by SLAM methods at lower resolution with the field of view fixed at 90\(^\circ\).
The authors then refine each of camera sequence using a structure-from-motion (SfM) pipeline, performing feature extraction, feature matching and global bundle adjustment successively.
Given the unawareness of global scene scale, the resulted camera poses of RealEstate10K are up to an arbitrary scale per clip.
For each frame the authors compute the 5-th percentile depth among all point depths from that frame’s camera. 
Computing this depth across all cameras in a video clip gives a set of near plane depths and the whole scene is scaled so that the 10-th percentile of this set of depths is 1.25m. 

While using RealEstate10K's scenes and camera trajectories during inference avoids scale issues within the dataset, challenges arise in more general cases. Specifically, when pairing out-of-domain images with either in-domain or out-of-domain trajectories, the inconsistencies between training and inference scales become evident. \textit{These inconsistencies make it impossible to generate realistic and controllable videos.}

The solution lies in reconstructing an absolute-scale scene for any given image. By leveraging metric depth predictor, we can reconstruct the absolute-scale 3D scene for the reference image. This absolute-scale scene bridges the gap between training and inference, enabling robust generalization to real-world applications. With this alignment, the model becomes capable of handling diverse combinations of images and trajectories, ensuring consistent and reliable performance across various scenarios.

\section{Training}

We choose DynamiCrafter~\cite{Xing2023} as our image-to-video (I2V) base model.
We trained proposed method on \(4\) publicly accessible variants of DynamiCrafter, namely \texttt{256}, \texttt{512}, \texttt{512\_interp} and \texttt{1024}.
We conduct ablation study on resolution 256\(\times\)256, due to the limitation of computing resource.
For resolution 256\(\times\)256, we train all models on \(\epsilon\)-prediction with effective batch size 64 for 50,000 steps, taking about 50 hours.
For resolution 512\(\times\)320 and 1024\(\times\)576, we train RealCam-I2V on \(v\)-prediction while enable \texttt{perframe\_ae} and \texttt{gradient\_checkpoint} to reduce peak GPU memory consumption.
We apply the Adam optimizer with a constant learning rate of \(1 \times 10^{-4}\) with mixed-precision fp16 and ZeRO-1.

For MotionCtrl~\cite{Wang2024Motionctrl} and CameraCtrl~\cite{He2024Cameractrl}, we reproduce all results on DynamiCrafter for fair comparison.
For CamI2V~\cite{zheng2024cami2v}, we implement hard mask epipolar attention and set 2 register tokens, aligned with the original paper. 
In quantitative comparison and ablation study, we set fixed text image CFG to 7.5 and camera CFG to 1.0.

\section{Depth Predictor}

We choose the metric depth version of Depth Anything V2~\cite{depth_anything_v2} as the metric depth predictor.
Compared to their basic versions, the authors fine-tune the pre-trained encoder on synthetic datasets for indoor and outdoor metric depth estimation.
The indoor model is capable of monocular metric depth estimation within a maximum depth of 20m.
We choose Large as the model size, which has 335.3M parameters, and the indoor version.
The scene scale of our model is aligned to the metric depth space of Depth Anything V2 Large Indoor, \ie absolute scene scale.

\begin{figure}[!t]
    \centering
    \vspace{-1mm}
    \includegraphics[width=\linewidth]{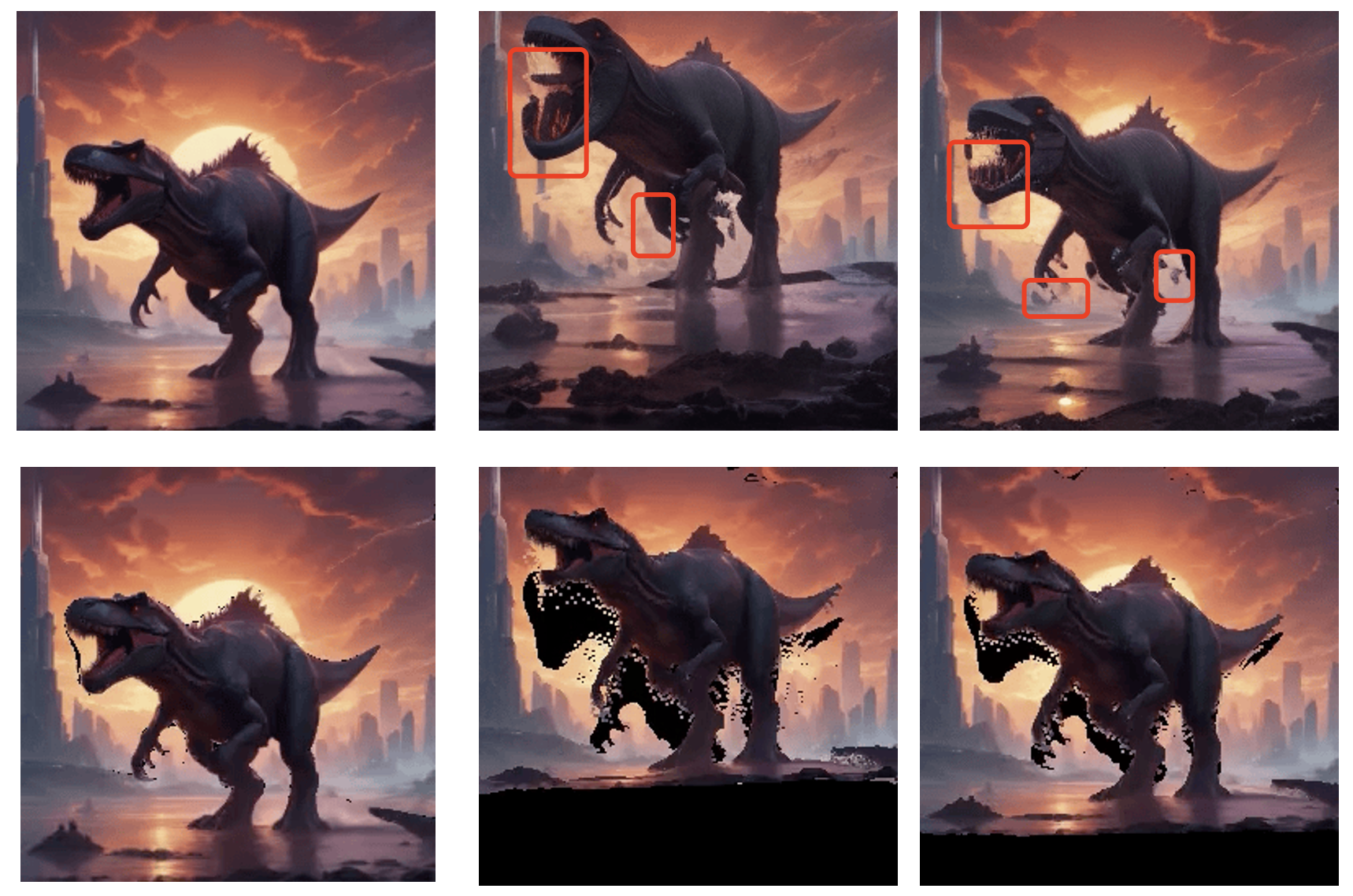}
    \vspace{-6mm}
    \caption{
    \textbf{Outlier filtering for noise shaping.}
    Undesired pixels will not be pasted in mistake for filtering kernel size $k\ge$3. } 
    \label{fig:rebuttal_kernel_size}
    \vspace{-5mm}
\end{figure}

\section{Noise Shaping and Parameter Sensitivity}

 The noise shaping mechanism is straightforward.
It overlays reference video features on model predictions at early diffusion steps. 
Fewer times of noise shaping means less prior imposed on layout, preserving more dynamics, and vice versa.
\cref{tab:rebuttal_noise_shaping} shows that effective control is typically achieved by selecting a threshold like $t_{\rm NS}=900$, similar to CFG. In practice, users only need to easily change between \{800, 900, 1000\}.
To enhance stability, we filter outliers of depth prediction on edge regions, as shown in~\cref{fig:rebuttal_kernel_size}.

\section{Robust Analysis of Depth Estimation, SfM, Base Model and Resolution}
\label{app:robustness}


Our framework demonstrates robustness across variations in depth predictors, SfM, base models, and high resolution.
The primary goal of our depth alignment is to establish a \textit{unified metric space}. 
This ensures that even if absolute depth values from predictors have errors, these errors are consistently propagated during alignment for both training and inference data. 
Consequently, there is no gap between training and inference despite inevitable errors from depth prediction and scale alignment.
The method can function even with less accurate metric depth predictors, though better predictors naturally enhance performance of our framework.

To further address this, we reproduce our method on RealCam-Vid~\cite{zheng2025realcam}, a high-resolution video dataset with dynamic scenes and metric-scale cameras.
We've successfully used both Depth Anything V2~\cite{depth_anything_v2} (metric indoor, 20m) and, in RealCam-Vid, Metric 3D v2~\cite{hu2024metric3dv2} (200m), showing strong adaptability to a different depth predictor, underscoring its robustness to depth estimation errors.
Our method works with both UNet-based (DynamiCrafter) and DiT-based (CogVideoX 1.5~\cite{yang2024cogvideox} in \cref{tab:rebuttal_realcam-vid}) backbones with higher resolution.
We also explored different SfM tools (e.g., GLOMAP~\cite{pan2024glomap} and MonST3R~\cite{zhang2024monst3r}) and found consistent performance.

\section{Real-time User Interface}

\cref{fig:one_round} in main paper illustrates our shift from slow, multi-round generation to an efficient one-round workflow, significantly enhancing usability. 
Users intuitively design camera paths by dragging within the reconstructed 3D scene, which is more direct than 2D or numerical input.
Users receive real-time preview feedback, rendered in parallel and streamed for immediate visualization, which is \textit{optional}, catering to different preferences. 
Our interactive preview is designed for nearly real-time feedback. Monocular depth estimation is efficient. For preview videos (e.g., 8 sampled keyframes), each frame is rendered \textit{independently} using modern engines (e.g., Open3D~\cite{Zhou2018open3d}) in a \textit{parallel} manner. This allows streaming playback, achieving real-time interaction. Further engineering optimization ensures a fluid user experience.

\section{Camera Keyframe Interpolation}

\begin{figure}[!t]
    \centering
    \vspace{-2mm}
    \includegraphics[width=0.8\linewidth]{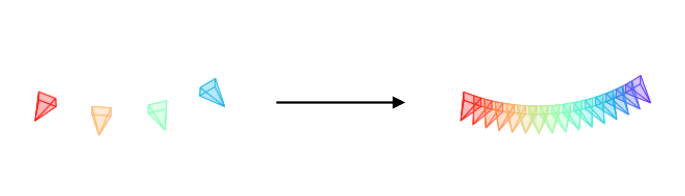}
    \vspace{-4mm}
    \caption{
        \textbf{Camera Trajectory Interpolation.}
        We interpolate camera keyframes given by user to dense trajectories.
    }
    \label{fig:camera_interp}
    \vspace{-4mm}
\end{figure}

In real-world applications, user-provided camera trajectories often consist of a limited number of keyframes (\eg, 4 keyframes). To ensure smooth and continuous motion across the trajectory while adhering to the user's input, we perform linear interpolation in SE(3) space to expand the trajectory to a higher number of frames (\eg, 16 interpolated frames), as shown in~\cref{fig:camera_interp}. This step ensures that our model generates consistent and visually coherent videos without compromising the accuracy of user-defined camera movements.

%% file: tables/rebuttal_noise_shaping.tex
\begin{table*}[ht!]
  \vspace{-1mm}
  \centering
    \resizebox{\linewidth}{!}{
    \renewcommand{\arraystretch}{1.0} 
    \begin{tabular}{c|cccccc|ccc}
    \toprule
      Noise Shaping   & Subject & Background & Motion & Dynamic & Aesthetic & Imaging & I2V & I2V & Camera \\
       Threshold  &   Consistency & Consistency & Smoothness & Degree & Quality & Quality &  Subject & Background & Motion \\
    \Xhline{0.75pt}
    \midrule
    /     & 90.99 & 96.23 & 97.36 & \textbf{46.75} & 58.37 & 62.91 & 94.73 & 93.44 & 85.85 \\
    900   & 93.96 & 97.58 & 97.66 & \underline{35.77} & 59.79 & 63.08 & 96.14 & 95.27 & 93.32     \\
    800   & \underline{95.02} & \underline{97.91} & \underline{97.79} & 26.42 & \underline{60.07} & \underline{63.33} & \underline{96.53} & \textbf{95.73} & \underline{95.15}  \\
    600   & \textbf{95.10} & \textbf{98.02} & \textbf{97.80} & 27.64 & \textbf{60.21} & \textbf{63.85} & \textbf{96.55} & \underline{95.72} & \textbf{96.72}   \\
            
    \bottomrule
  \end{tabular}
  }
  \vspace{-2.5mm}
  \caption{
    \textbf{Evaluation results of noise shaping threshold $t_{\rm NS}$.}
  For noise level $t \in [0, 1000]$, the threshold $t_{\rm NS}$ denotes that we apply noise shaping only when $t>t_{\rm NS}$ in early denoising process.}
  \label{tab:rebuttal_noise_shaping}
  \vspace{-1mm}
\end{table*}

%% file: tables/rebuttal_realcam-vid.tex
\begin{table*}[ht!]
  \vspace{-1mm}
  \centering
    \resizebox{\linewidth}{!}{
    \renewcommand{\arraystretch}{1.0} 
    \begin{tabular}{c|cccccc|ccc}
    \toprule
      \multirow{2}{*}{Model} & Subject & Background & Motion & Dynamic & Aesthetic & Imaging & I2V & I2V & Camera \\
        &   Consistency & Consistency & Smoothness & Degree & Quality & Quality &  Subject & Background & Motion \\
    \Xhline{0.75pt}
    \midrule
    CogVideoX 1.5~\cite{yang2024cogvideox}    & 91.80 & 94.66 & 97.07 & 40.98 & 62.29 & 70.21  & 96.46 & 95.50 & 39.71 \\
    \rowcolor{gray!15}
    \textit{w.} RealCam-I2V (Ours)     & \textbf{97.81} & \textbf{98.41} & \textbf{99.33} & \textbf{44.31} & \textbf{64.31} & \textbf{70.70} & \textbf{99.03} & \textbf{99.40} & \textbf{91.35} \\
            
    \bottomrule
  \end{tabular}
  }
  \vspace{-2.5mm}
  \caption{
      \textbf{VBench-I2V results of RealCam-I2V trained on RealCam-Vid~\cite{zheng2025realcam}}, with scene dynamics and large camera movements (near 360$^\circ$).
      The improvement in metrics can be attributed to the additional information from diverse camera traces in dynamic scenes. 
      Noise shaping significantly improves the consistency and quality.
  }
  \label{tab:rebuttal_realcam-vid}
  \vspace{-4mm}
\end{table*}